\documentclass[journal=jpcl,manuscript=article]{achemso}

\usepackage[version=3]{mhchem} 
\usepackage{booktabs}
\usepackage{xcolor}

\newcommand{\bftab}{\fontseries{b}\selectfont}

\author{Rishikesh Magar}
\author{Yuyang Wang}
\author{Amir Barati Farimani}

\affiliation[Carnegie Mellon University]
{Department of Mechanical Engineering, Carnegie Mellon University, Pittsburgh, PA 15213}
\email{barati@cmu.edu}

\title[An \textsf{achemso} demo]
  {Crystal Twins: Self-supervised Learning for Crystalline Material Property Prediction}

\abbreviations{IR,NMR,UV}

\begin{document}







\begin{abstract}
Machine learning (ML) models have been widely successful in the prediction of material properties. However, large labeled datasets required for training accurate ML models are elusive and computationally expensive to generate. Recent advances in Self-Supervised Learning (SSL) frameworks capable of training ML models on unlabeled data have mitigated this problem and demonstrated superior performance in computer vision and natural language processing tasks. Drawing inspiration from the developments in SSL, we introduce Crystal Twins (CT): an SSL method for crystalline materials property prediction. Using a large unlabeled dataset, we pre-train a Graph Neural Network (GNN) by applying the redundancy reduction principle to the graph latent embeddings of augmented instances obtained from the same crystalline system. By sharing the pre-trained weights when fine-tuning the GNN for  regression tasks, we significantly improve the performance for 7 challenging material property prediction benchmarks. 
\end{abstract}


\section{Introduction}

Machine Learning (ML) based predictive models have made rapid strides in computational chemistry due to their efficiency and performance. 
Characterized by their computational efficiency and accuracy, these methods are capable of faster high-throughput screening compared to classical physics models \cite{schmidt2019recent,keith2021combining}. This capability has roots in both novel learning algorithms and improved hardware. Even though ML models can offer faster predictions, the accuracy of these models is highly correlated with the availability of clean labeled data. \cite{najafabadi2015deep}. In general, it is difficult to develop accurate and robust ML models without sufficiently large labeled data \cite{bengio2021deep}. Moreover, the acquisition of labeled data is expensive as it involves performing Density Functional Theory (DFT) simulations or experiments to characterize materials \cite{schleder2019dft,chen2020machine}. On the other hand, gigantic databases containing structures and compositions of materials without labels (property) are available. These databases can not be used in supervised learning tasks due to the lack of labels. Given the availability of large unlabeled datasets, some interesting questions can be asked 1.) can we develop more efficient ML models that are capable of learning the underlying structural chemistry from unlabeled data and 2.) can these models be used to make the supervised learning tasks more accurate?

In this work, we aim to address these questions by leveraging Self-Supervised Learning (SSL) for material property prediction. Unlike supervised learning which uses labels for supervision, SSL makes use of the large unlabeled data for supervision to learn robust and generalizable representations that can be used for various tasks. Recently, SSL frameworks such as SimCLR\cite{chen2020simple} , Barlow Twins \cite{zbontar2021barlow} , BYOL\cite{grill2020bootstrap} , SwAV \cite{caron2020unsupervised} , momentum contrastive learning \cite{he2020momentum} , SimSiam \cite{chen2021exploring} , Albert \cite{lan2019albert} , and self-supervised dialogue learning \cite{wu2019self} have been successfully applied to computer vision and natural language processing tasks. The success of these SSL methods has inspired many works in molecular ML, leading to the development of highly accurate frameworks such as MolCLR \cite{wang2022molecular} , dual view molecule pre-training \cite{zhu2021dualview} , 3D Infomax \cite{stark20213d} , and numerous other popular works\cite{liu2019n,rong2020self,hu2019strategies,li2021effective,chithrananda2020chemberta,rong2020grover,zhang2021motif, wang2022improving}. 
\begin{figure}[t!]
    \centering
    \includegraphics[width=\textwidth, keepaspectratio=true]{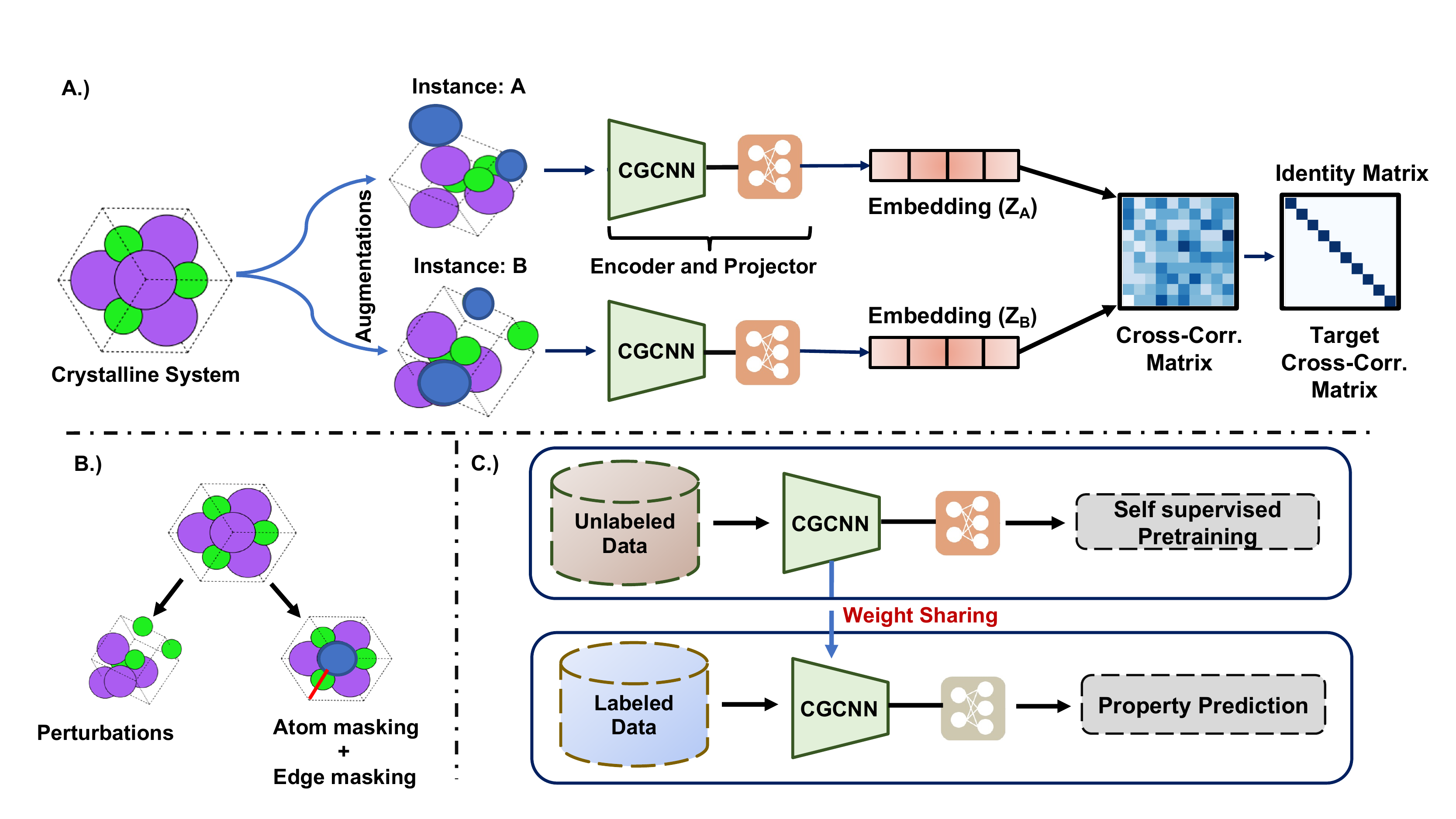}
    \caption{Overview of the Crystal Twins (CT) framework. A.) The Crystal Twins framework takes  the structural file (CIF) as the input and then augments the structure to create two different augmented instances. Each instance is passed to the CGCNN graph encoder followed by a projector to generate embedding. The pre-training objective aims to maximize the cross-correlation between the two embeddings. B.) To create augmented instances, three augmentation techniques are used in this work: random perturbations, atom masking, and edge masking. C.) In the pre-training stage we trained using SSL. In the fine-tuning stage,the pre-trained weights are shared with the encoder(CGCNN) which is connected MLP head to predict the material property.}
    \label{fig:CT_pipeline}
\end{figure}

 Most of the promising works developed for material property prediction tasks are using graph neural networks (GNN). GNNs consider non-Euclidean topology to construct a graph representation that can be learned and modified according to the task \cite{lecun2015deep, wu2020comprehensive,kipf2017semisupervised}. GNNs developed for material property prediction include CGCNN\cite{xie2018crystal} , OGCNN\cite{karamad2020orbital} , SchNet\cite{schutt2018schnet} , MegNet\cite{chen2019graph}, and other models \cite{2020GATGNN,klicpera2020directional,klicpera2020fast,palizhati2019toward,back2019convolutional,gilmer2017neural,unke2019physnet,gu2020practical,jha2018elemnet,dunn2020benchmarking,choudhary2021atomistic}. Developments have also been made in tasks such as material structure generation and prediction \cite{moosavi2020role,ryan2018crystal,liang2020cryspnet,long2021constrained,kim2020generative,xie2021crystal} as well as identifying new materials with specific properties\cite{yao2021inverse}. Despite progress being made in developing self-supervised ML architectures in the molecular ML, there is a noticeable lack of research works implementing such techniques for the periodic crystalline systems property prediction. In this work, we introduce Crystal Twins (CT): an SSL framework for crystalline material property prediction with GNNs (Figure~\ref{fig:CT_pipeline}). In pre-training, the CT framework does not make use of any labeled data to learn crystalline representations, instead, it trains ML models in a self-supervised manner. 
In the CT framework, we use the CGCNN\cite{xie2018crystal} as the encoder to learn expressive representations of crystalline systems (Figure~\ref{fig:CT_pipeline}a). The GNN encoder (i.e., CGCNN) generates representations of two augmented instances from the same crystal and the objective of CT pre-training is to make the cross-correlation matrix of the two embeddings as close as possible to the identity matrix (Figure~\ref{fig:CT_pipeline}a). To create augmented instances, we introduce the combination of three different augmentation techniques: random perturbations, atom masking, and edge masking (Figure~\ref{fig:CT_pipeline}b). The representations learned by the encoder are later used for downstream material property prediction tasks in the fine-tuning stage (Figure~\ref{fig:CT_pipeline}c). In the pre-training stage,  the CT model learns representations without any labeled data. Using the pre-trained weights to initialize the GNN encoder for fine-tuning, CT demonstrates superior prediction performances on 7 challenging datasets. We also compare the performance of the CT model with other competitive supervised learning baselines.  


\section{Methods}
In this section, we describe the components of the CT framework (Figure~\ref{fig:CT_pipeline}). 
In general, SSL frameworks employ correlations in the input itself to learn robust and generalizable representations from unlabeled data. \cite{hadsell2006dimensionality}. 
In the CT framework, the goal during pre-training is to force the empirical cross-correlation matrix created from the encoder embeddings  two different augmentations generated by the same crystal towards the identity matrix. All the elements in the cross-correlation matrix lie between -1 and 1, with 1 representing maximum correlation. Intuitively, since the embeddings are generated from augmentations of the same crystalline system, the cross-correlation matrix must be close to the identity matrix. Using such an objective during pre-training allows the graph encoder to learn robust representations. To create the augmented instances, we use augmentation techniques, including atom masking, edge masking, and random perturbation. The embeddings for the augmented instances of the crystalline system are generated via the CGCNN graph encoder. To pre-train the CGCNN in a self-supervised manner, we use the Barlow Twins loss function. The weights of the pre-trained self-supervised model are used to initialize the graph encoder model during the fine-tuning stage for material property prediction.

Most of the successful deep learning approaches for crystalline material property prediction are based on GNNs because of their ability to capture structural geometry and chemistry. In a crystal graph ($G$), we consider the atoms as the nodes ($V$), and interactions between them are modeled via edges ($E$). In general, GNNs aggregate information from the neighborhood of the node to construct embeddings that are updated iteratively. The update for the GNN can be described as in Equation~\ref{eq:aggregate}
\begin{align}
\begin{split}
    \pmb h_v^{(k)} &= \text{COMBINE}^{(k)} \bigg(\pmb h_v^{(k-1)}, \text{AGGREGATE}^{(k)} \Big( \{\pmb h_u^{(k-1)} \vert u \in \mathcal{N}(v)\} \Big) \bigg),
\end{split}
\label{eq:aggregate}
\end{align}
where $\pmb h_v^{(k)}$ is the feature of the node $v$ at the $k$-th layer and $\pmb h_v^{(0)}$ is initialized by node feature $\pmb x_v$. $\mathcal{N}(v)$ denotes the set of all the neighbors of node $v$. $\pmb{a}_v^{(k)}$ is the output from the aggregation operation at the $k^{th}$ layer. The aggregation operation collects the features of neighboring nodes and the combination operation combines the original node feature with the aggregated features. To extract the feature of the entire crystal system, $\pmb h_G$, readout operation integrates all the node features among the graph $G$ as given in Equation~\ref{eq:readout}:
\begin{equation}
    \pmb h_G = \text{READOUT} \Big( \{\pmb h_v^{(k)} \vert v \in G\} \Big).
    \label{eq:readout}
\end{equation}
The readout operations such as summation, averaging, and max pooling are most commonly used\cite{xu2018how}. 

In this work, we implement the CGCNN\cite{xie2018crystal} architecture as the GNN encoder. We choose CGCNN because of its competitive performance and computational efficiency when compared to other GNN baselines. To encode crystal features and obtain an embedding, we use mean pooling to generate a latent representation with the dimension of 64. To expand the latent embedding and have a better combination of encoded features, we use a projector with 2 MLP layers that generates a final embedding of size 128. We use this embedding to generate the cross-correlation matrix for applying the Barlow Twins loss.

In the pre-training stage, we use the Barlow Twins loss function to learn graph representations from crystals. This loss is based on the redundancy reduction principle proposed by neuroscientist H. Barlow \cite{barlow2001redundancy,barlow1961possible} and  was introduced to SSL by Zbontar et al.\cite{zbontar2021barlow}. We use the Barlow Twins loss function in CT because of its high performance and ease of implementation. The Barlow Twins loss function is applied to the cross-correlation matrix created from encoder-generated embeddings of the two different augmentations of the same crystalline system. The Barlow Twins loss function is represented by Equation~\ref{eq1},
\begin{equation}
    L_{BT} \overset{\Delta}{=} \sum_i(1-C_{ii})^2 + \lambda\sum_i\sum_{j\neq i} C_{ij}^2,
    \label{eq1}
\end{equation}
where $C$  is the cross-correlation matrix of embeddings from two augmented instances, the cross correlation matrix is given by Equation ~\ref{eq2}. The $\lambda$ used in this work is $0.0051$ same as the original paper. 
\begin{equation}
    C_{ij} \overset{\Delta}{=} \frac{\sum_b Z_{b,i}^A Z_{b,j}^B}{\sqrt{(Z_{b,i}^A)^2}{\sqrt{(Z_{b,j}^B)^2}}} 
    \label{eq2}
\end{equation}
where $b$ is the index of the batch and $i,j$ index the vector dimensions of the projector output ($Z^A$ and $Z^B$), for both the augmented instances $A$ and $B$ from the same crystalline material.

To generate self-supervised learning representations, we need to construct different augmentations of the crystalline system. Inspired by AugLiChem,\cite{magar2021auglichem} we devise three different augmentation techniques (Figure~\ref{fig:CT_pipeline}b), namely, random perturbation, atom masking and edge masking. The random perturbation augmentation perturbs all the atoms in the crystalline system by a distance of 0.05 \AA. For atom masking, we mask 10\% of the atoms in the crystal randomly, similarly for edge masking we randomly mask 10\% of the edge features between two neighboring atoms. More details on atom masking and edge masking are provided in the supplementary information (Figure S1). These augmentations are applied to the crystalline systems and two augmented instances are generated for pre-training.

In the pre-training stage, the embedding dimension of the CGCNN encoder is set to 128. We use the Adam optimizer\cite{kingma2014adam} with a learning rate of $0.00001$ and a batch size of $64$ and pre-train the model for $15$ epochs. The other hyperparameters for the CGCNN model are kept the same as in the original paper. The train/validation ratio for pre-training data is 95\%/5\%. For pre-training, we combine the datasets from the Matminer database \cite{ward2018matminer} and the hypothetical Metal-Organic Framework dataset\cite{wilmer2012large}, aggregating a total of 428,275 samples. The labels in the datasets are not used during CT pre-training. In the fine-tuning stage, we add a randomly initialized MLP head with two fully connected layers to generate the final property prediction. The model in the fine-tuning stage is trained for 200 epochs. Similarly, the results for supervised CGCNN are also reported after training the model for 200 epochs. For fine-tuning we use the Adam optimizer with a learning rate of $0.001$ and batch size 128. 

\section{Results}
To comprehensively evaluate the performance of the CT framework, we test its performance on seven challenging regression benchmark datasets. The properties we evaluate include formation energy, band gap, Fermi energy, and shear modulus for different materials. An overview of the dataset used for benchmarking the performance of the CT framework is shown in Table~\ref{tb:benchmark} . A detailed description of these datasets is available in the supplementary information.

\begin{table}[htb!]
  \centering
  \small
  \footnotesize
  \resizebox{\textwidth}{!}{\begin{tabular}{l|llllll}
    \toprule
    Dataset & \# Crystals & Property  & Unit \\
    \midrule
    HOIP & 1333 & Band gap & \textit{eV}\\
    Lanthanides & 4166 & Formation energy & \textit{eV/atom}\\
    GVRH & 10987 & Shear Modulus &  \textit{$\log_{10}VRH$}\\
    Perovskites & 18928 & Formation Energy  & \textit{eV/atom}\\
    
    Fermi Energy & 26447 & Fermi energy & \textit{eV}\\
    
    Formation Energy (FE) & 26078 & Formation energy  & \textit{eV/atom} \\
    Band Gap (BG) & 26709 & Band gap  & \textit{eV}\\
   \bottomrule
  \end{tabular}}
  \caption{Overview of the datasets used for benchmarking the performance of the CT framework}
  \label{tb:benchmark}
\end{table}
As we pre-trained the model with CGCNN encoder, the comparison with the supervised CGCNN model is the most direct and fair, and it offers insights into how self-supervised learning methods can help in predicting the crystalline material properties with a high degree of accuracy. We also compare the performance of the CT framework with other popular supervised GNN models, i.e., GATGNN\cite{2020GATGNN}, GIN\cite{hu2019strategies}, and OGCNN\cite{karamad2020orbital}. All the models used for comparison are trained with the same hyperparameters as suggested in their publicly available codes. The train/validation/test split for the datasets is set to $0.6/0.2/0.2$ following previous standard benchmarking protocols. The test Mean Absolute Errors (MAEs) for the supervised training baselines and the CT framework are shown in Table~\ref{tb:comparison}. The detailed hyperparameters used for supervised models are listed in the supplementary information (Table S2).

\begin{table}[htb!]
  \centering
  \small
  \footnotesize
  \resizebox{\textwidth}{!}{\begin{tabular}{l|lllllll}
    \toprule
    Dataset & HOIP\cite{kim2017hybrid} & Lanthanides\cite{PhamOrbital} & GVRH\cite{ward2018matminer,deJong2015} & Perovskites\cite{Castelli} & Fermi Energy\cite{materialsproject} & FE\cite{materialsproject} & BG\cite{materialsproject}\\
  \# Crystals & 1333 & 4166 & 10987 & 18928 & 26447 & 26078 & 26709  \\
    \midrule
    GIN \cite{hu2019strategies} & 0.666$\pm$0.123 & 0.197$\pm$0.038 & 0.133$\pm$0.007 & 0.380$\pm$0.008 & 0.605$\pm$0.015 & 0.109$\pm$0.007 & 0.601$\pm$0.038  \\
    CGCNN \cite{xie2018crystal} & 0.170$\pm$0.013 & 0.080$\pm$0.003 & 0.092$\pm$0.001 & 0.057$\pm$0.002 & \bftab{0.400$\pm$0.003} & 0.040$\pm$0.001 & 0.369$\pm$0.003  \\
    GATGNN \cite{2020GATGNN} & 0.169$\pm$0.016 & 0.097$\pm$0.004 & 0.091$\pm$0.000 & 0.071$\pm$0.002 & 0.428$\pm$0.007 & 0.050$\pm$0.004 & 0.376$\pm$0.007  \\
    OGCNN \cite{karamad2020orbital} & 0.164$\pm$0.013 & 0.072$\pm$0.002 & 0.089$\pm$0.001 & 0.056$\pm$0.001 &0.446$\pm$0.018 & 0.035$\pm$0.001 & 0.353$\pm$0.008 \\
    \midrule
    CT & \bftab{0.153$\pm$0.003} & \bftab{0.058$\pm$0.001} & \bftab{0.087$\pm$0.001} & \bftab{0.046$\pm$0.001} & \bftab{0.399$\pm$0.004} & \bftab{0.025$\pm$0.001} & \bftab{0.328$\pm$0.002} \\
    \bottomrule
  \end{tabular}}
  \caption{Mean and standard deviation of test MAE of Crystal Twins (CT) in comparison to the supervised baselines on regression benchmarks.}
  \label{tb:comparison}
\end{table}

It is observed that the CT model outperforms all supervised learning baselines on all the 7 regression tasks. We would like to note that the performance improvements (Table S1) achieved by the CT over the baseline supervised CGCNN model is non-trivial. We observed an average improvement of 15.55\% improvement for CT over supervised-learning CGCNN. The results in Table~\ref{tb:comparison} clearly demonstrate the merit of using self-supervised learning frameworks for periodic crystal property prediction.



\begin{figure}[htb!]
    \centering
    \includegraphics[width=\textwidth, keepaspectratio=true]{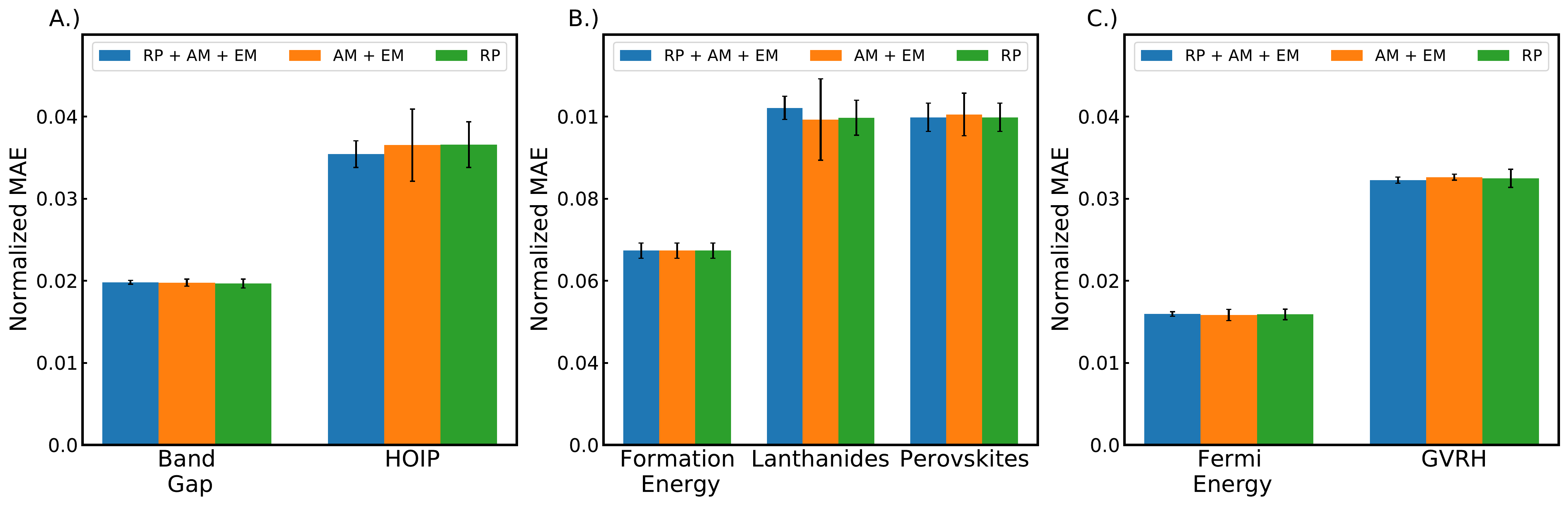}
    \caption{Ablation study of three augmentation techniques, random perturbation (RP), atom masking (AM), and edge masking (EM), for CT framework. A.) Evaluating the effect of different augmentation techniques in Band Gap and HOIP dataset where the label is band gap. B.) Evaluating the effect of augmentation techniques on the FE, Lanthanides, and Perovskites datasets for which the label is formation energy. C.) Evaluating the effect of different augmentation strategies on the Fermi energy  and log10 VRH - shear modulus of the structures prediction.}
    \label{fig:CT_ablation}
\end{figure}

To compare the effectiveness of the different augmentations techniques, we pre-train three CT models, 1) using only random perturbation augmentations (RP), 2) using only atom masking and edge masking augmentations (AM+EM), 3) using all three random perturbation, atom masking, and edge masking augmentations (RP+AM+EM). We report the MAE of the model on different fine-tuning tasks to determine the effective augmentation techniques (Figure~\ref{fig:CT_ablation}).

The performance of AM+EM augmentation is better than RP for perovskites, BG and GVRH datasets , whereas RP augmentation has better performance than AM+EM for Fermi energy, lanthanides, and HOIP datasets. For FE dataset the performance of both RP and AM+EM augmentation techniques is the same. It must be noted that the performance of models trained with different augmentation techniques is almost identical, making it difficult to conclusively ascertain which augmentation technique is better. Moreover, we also observe that the effectiveness of the augmentation techniques is dataset dependent. We would also like to note that the standard deviation of MAE is always lower when using the pre-trained model with all augmentation techniques. Therefore, using a combination of all three augmentation techniques is most effective.

 

\begin{figure}[htb!]
    \centering
    \includegraphics[width=\textwidth, keepaspectratio=true]{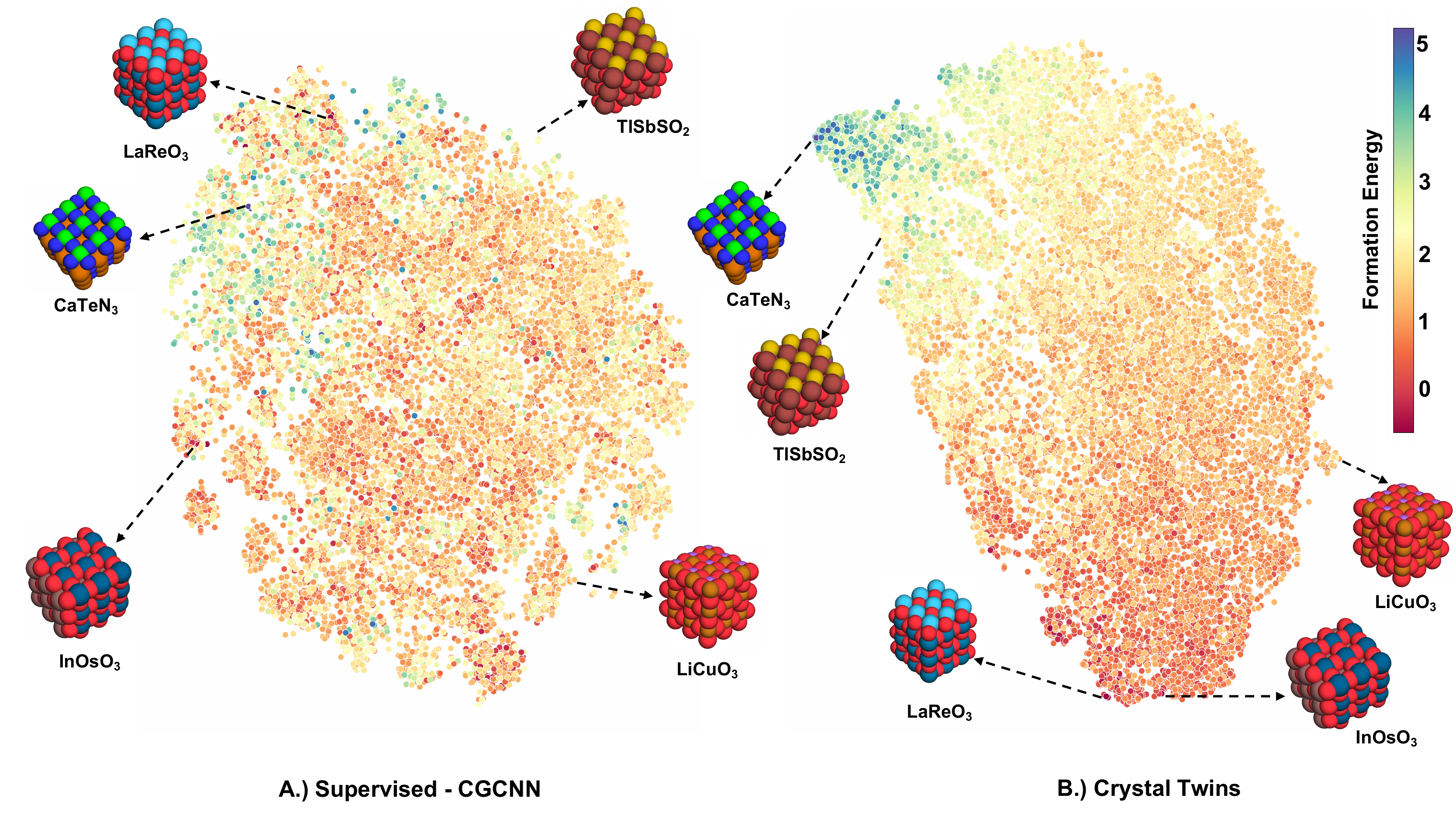}
    \caption{Visualizing the embeddings space for the Perovskites dataset using t-SNE. Every point on the t-SNE plot is colored corresponding to the formation energy of the crystalline system. A.) The t-SNE plot for the embedding was generated from the supervised CGCNN model B.) The t-SNE plot for the embedding was generated from the graph encoder of CT model after fine-tuning. }
    \label{fig:t-SNE}
\end{figure}

To understand the CT representations, we visualize the representations from the  pre-trained and fine-tuned CT framework in comparison to the supervised CGCNN in 2D using t-SNE\cite{van2008visualizing}. The t-SNE representation maps the embedding based on the similarity in the 2D space. The comparison between the representations of the supervised CGCNN model and the CT model for the perovskites dataset is shown in Figure~\ref{fig:t-SNE}. Each point is colored by the formation energy of perovskites which is the label that the model is trained on in the fine-tuning stage.  

We observe that the t-SNE projection from the CT model has a better separation, the crystalline materials with higher formation energy are clustered at the top left of the t-SNE projection plot (Figure~\ref{fig:t-SNE}B) when compared to the supervised CGCNN (Figure~\ref{fig:t-SNE}A). Similarly the materials with lower formation energy are clustered at the bottom of the graph(Figure~\ref{fig:t-SNE}B).  For example, perovskites $InOsO_3$ and $LaReO_3$ with relatively lower formation energies of -0.58 and -0.64 $eV/atom$ respectively are clustered closely together in t-SNE projection from CT than from supervised CGCNN.  This demonstrates the generalizability of the representations learned by the CT model when compared to supervised learning. Such representation learnt from the CT framework can also be used to characterize and understand the large chemical space of materials.

\section{Conclusion}
In this work, we develop Crystal Twins (CT), an SSL framework for crystalline material property prediction. The CT model achieves a superior performance compared to other competitive supervised learning baselines. The CT framework demonstrates high generalizability and robustness by learning representations that can be used to predict a variety of properties like formation energy, band gap, and Fermi energy of different crystalline materials. The CT framework has been trained on significantly less data when compared SSL models in other domains like molecular machine learning, computer vision and natural language processing. In general, SSL models are known to demonstrate a better performance with larger unlabeled data as it allows them to learn more generalizable representations. We expect the CT model to demonstrate a superior performance with larger training data when compared to our current results. The representations learned by the CT model are of great promise and can open up avenues for exciting research in understanding the chemical space and designing materials with desired properties.        

\begin{acknowledgement}
The work has been supported by the start-up fund provided by Department of Mechanical Engineering at Carnegie Mellon University. The authors would like to thank Prakarsh Yadav and Alison Bartsch for their comments on the manuscript. 

\end{acknowledgement}

\section*{Data and Code Availability}

All data used in this work is publicly available. Please contact the corresponding author for more details about code and data.


\newpage
\bibliography{achemso-demo}

\newpage







\end{document}


\newpage
\section{Augmentation Techniques}
We introduce three augmentation techniques in this work for creating augmented instances. The intuition behind SSL strategies is that if augmented instances are created from same crystalline materials there should be maximal agreement between the embeddings generated for the graph encoder. For crystalline materials, we used three augmentation techniques random perturbation, atom masking and edge masking. The random perturbation augmentation technique is shown to effective in Auglichem\cite{magar2021auglichem} and previous works done by You et al. have shown the effectiveness of masking techniques\cite{you2020graph}.

\subsection{Random Perturbation}
We use the random augmentation technique as introduced in Auglichem\cite{magar2021auglichem}. In the random perturbation augmentation technique all the augmentation atoms in the crystalline system are perturbed between 0 to 0.05\AA. This augmentation enables breaking of the symmetries in a crystalline material and updates the neighbor list based on which the graph is constructed.

\subsection{Atom Masking}
\begin{figure}[htb!]
    \centering
    \includegraphics[width=\textwidth, keepaspectratio=true]{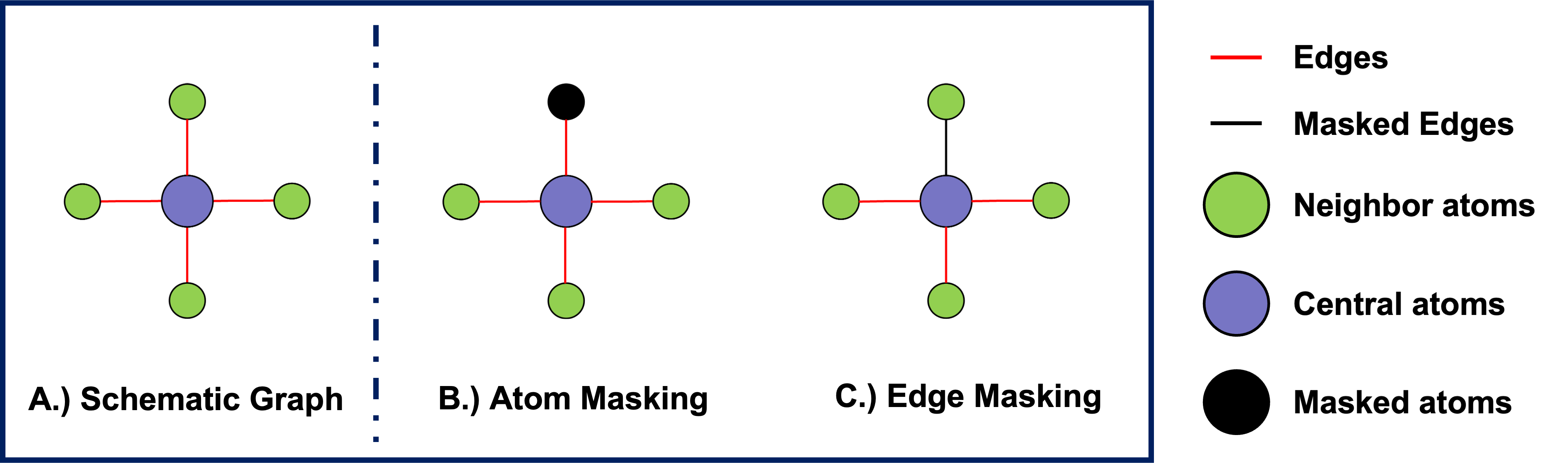}
    \caption{a.) Schematic Graph of a crystalline material. The central atom of the graph is denoted in the color purple. The neighbor atoms are denoted in green. The edges between the central atom and neighbor atoms are marked in red. b.)Atom Masking strategy, the masked atom is shown in black. c.) Edge masking strategy, the masked edge is shown in black. }
    \label{fig:AM_EM}
\end{figure}

The schematic graph for the crystalline material is shown in Figure~\ref{fig:AM_EM}a, the graph representation is constructed is based on distances of the nearest neighbors to the central atom. The node features encode the chemical properties of the atoms and the edge features encodes the relationship between the central atom and the neighbor atoms. In the atom masking technique, 10\% of the atoms in the crystalline system are masked(Figure~\ref{fig:AM_EM}b). In the masking operation all the node features corresponding to the masked atom are set to zero. The graph is constructed with a placeholder atom whose node features are zero. Using such an augmentation technique helps improve the robustness of the representations learnt by the model in the pre-training stage. 

\subsection{Edge Masking}
In the edge masking technique, 10\% of the edges between the central atom and its nearest neighbors are masked. Similar to the atom masking operation, in edge masking all the edge features corresponding to the masked edge are set to zero and graph is constructed. 

\newpage
\section{Dataset Details}
In this work we use seven challenging publicly available datasets to benchmark the performance of machine learning models. The brief description of the datasets are as follows.

\subsection{Pre-training Dataset}

For pre-training, we use the datasets from Matminer database \cite{ward2018matminer} and the hypothetical Metal Organic Framework dataset\cite{wilmer2012large}. The hMOF database contributed 275,907 data points and the matminer database contributed 152,368 samples. In total for pretraining we used 428,275 samples.

\subsection{Hybrid Organic Inorganic Perovskites (HOIP) dataset}
The HOIPs dataset has been made publicly available by Kim et al. \cite{kim2017hybrid}. The dataset consists of HOIP structural file and the perform density functional theory to calculate a variety of properties such as band gap, relative energies etc. In this work, we predict the band gap for HOIP. 
\subsection{Lanthanides}
The Lanthanides dataset consists of structural files formation energies of 4133 lanthanide element based alloys.\cite{PhamOrbital} 

\subsection{GVRH}
The dataset has been taken from the matminer repository\cite{Dunn2020,deJong2015}. The dataset consists of 10987 structural files and the task is to predict the DFT calculated {$\log_{10}VRH$} - shear modulus of the structures.

\subsection{Perovskites}
This dataset contains 18928 structural files of perovskites type materials. The perovskites in this data take the form $ABX_3$, where $A$ and $B$ are cations and $X$ is the halide ion. The perovskites dataset was by developed by Castelli et al. \cite{Castelli} and we predict the formation energy of the structures.

\subsection{Fermi Energy}
The Fermi Energy dataset consists of 26447 structural files from the Materials Project \cite{materialsproject}. In the dataset the task is to predict the DFT calculated Fermi energies.

\subsection{Formation Energy}
The Formation Energy(FE) dataset consists of 26078 structural files from the Materials Project \cite{materialsproject}. In the dataset the task is to predict the DFT calculated Formation Energy.

\subsection{Band Gap}
The Band Gap(BG) dataset consists of 26709 structural files from the Materials Project \cite{materialsproject}. In the dataset the task is to predict the DFT calculated Band Gap. \\

\newpage
\section{Improvement of CT over Supervised learning baselines}
\begin{table}[htb!]
  \centering
  \small
  \footnotesize
  \resizebox{\textwidth}{!}{\begin{tabular}{l|lllllll|l}
    \toprule
    Dataset & HOIP & Lanths & GVRH & Perovskites & Fermi Energy & FE & BG & Average \\
    \# Crystals & 1333 & 4166 & 10987 & 18928 & 26447 & 26078 & 26709 & Improvement \\
    \midrule
    \% Improvement over CGCNN \cite{xie2018crystal} & 9.82 & 27.5 & 4.71 & 18.71 & 0.33 & 36.88 & 10.93 & 15.55  \\
    \% Improvement over GATGNN \cite{2020GATGNN} & 9.46 & 40.67 & 3.66 & 34.43 & 6.77 & 50.32 & 12.78 & 22.78 \\
    \% Improvement over GIN \cite{hu2019strategies} & 77.02 & 70.55 & 34.08 & 87.80 & 34.04 & 76.45 & 45.31 & 60.75  \\
    \% Improvement over OGCNN \cite{karamad2020orbital} & 6.89 & 19.81 & 1.49 & 17.75 & 10.60 & 28.03 & 6.98 & 13.07\\

    \bottomrule
  \end{tabular}}
  
  \caption{\% Improvement of the Crystal Twins model over the  baseline supervised learning models on all the datasets.}
  \label{tb:Improvement}
\end{table}

\newpage
\section{Hyperparameters for the baseline supervised models}
The hyperparamters were kept the same as the optimized settings in the publicly available code  repositories and the manuscripts describing the models. In Table~\ref{tb:Hyperparam}, we describe the hyperparameters used when training the models. For additional details about other hyperparameters, please refer to the manuscripts accompanying the models

\begin{table}[htb!]
  \centering
  \resizebox{\textwidth}{!}{\begin{tabular}{l|lllllll}
    \toprule
    Hyperparamters & Batch Size & Num. GCN Layers & Learning Rate & Optimizer & Epochs &  Train/Val/Test \\
    \midrule
    CGCNN \cite{xie2018crystal}& 256 & 3 & 0.01 & SGD & 200 & 0.6/0.2/0.2   \\
    GATGNN \cite{2020GATGNN} & 256 & 3 & 0.005 & AdamW & 200 &  0.6/0.2/0.2    \\
    GIN \cite{hu2019strategies} & 256 & 5 & 0.001 & Adam & 100 &  0.6/0.2/0.2 \\
    OGCNN\cite{karamad2020orbital} & 32 & 3 & 0.0005 & SGD & 100 & 0.6/0.2/0.2  \\

    \bottomrule
  \end{tabular}}
  \caption{Hyperparameters for baseline supervised learning models and finetuning of CT model.}
  \label{tb:Hyperparam}
\end{table}

\bibliography{achemso-demo}